\documentclass{article}

\usepackage[preprint]{neurips_2026}

\usepackage[utf8]{inputenc} %
\usepackage[T1]{fontenc}    %
\usepackage{hyperref}       %
\usepackage{url}            %
\usepackage{booktabs}       %
\usepackage{amsfonts}       %
\usepackage{nicefrac}       %
\usepackage{microtype}      %
\usepackage{xcolor}         %

\usepackage{graphicx}
\usepackage{algorithm}
\usepackage{algorithmic}
\usepackage{amsmath}
\usepackage{wrapfig}
\usepackage{subcaption}
\usepackage{makecell}
\usepackage{amsthm}

\newcommand{\method}{\textsc{Orbit}}
\title{\method: Preserving Foundational Language Capabilities in GenRetrieval via Origin-Regulated Merging}

\author{%
Neha Verma\thanks{Equal contribution. Work done while NV was at Google.} \\
  Johns Hopkins University \\
  \texttt{nverma7@jhu.edu} \\
  \And
  Nikhil Mehta$^{*}$\\
  Google DeepMind \\
\texttt{nikhilmehta.dce@gmail.com} \\
  \AND
  Shao-Chuan Wang \\
  Google DeepMind \\
  \And
  Naijing Zhang \\
  Google \\
  \And
  Alicia Tsai \\
  Google DeepMind \\
  \And
  Aniruddh Nath \\
  Google \\
  \And
  Li Wei \\
  Google \\
  \And
  Lukasz Heldt \\
  Google \\
  \And
  Lichan Hong \\
  Google DeepMind \\
  \And
  Ed Chi \\
  Google DeepMind \\
  \And
  Xinyang Yi \\
  Google DeepMind\\
}

\begin{document}

\maketitle
\begin{abstract}
Despite the rapid advancements in large language model (LLM) development, fine-tuning them for specific tasks often results in the catastrophic forgetting of their general, language-based reasoning abilities. 
This work investigates and addresses this challenge in the context of the Generative Retrieval (GenRetrieval) task. 
During GenRetrieval fine-tuning, we find this forgetting occurs rapidly and correlates with the distance between the fine-tuned and original model parameters. 
Given these observations, we propose \method{}, a novel approach that actively tracks the distance between fine-tuned and initial model weights, and 
uses a weight averaging strategy to constrain model drift during GenRetrieval fine-tuning when this inter-model distance exceeds a maximum threshold. 
Our results show that \method{} retains substantial text and retrieval performance by outperforming both common continual learning baselines and related regularization methods that also employ weight averaging.
\end{abstract}

\section{Introduction}
\label{introduction}

The Generative Retrieval (GenRetrieval) paradigm~\citep{rajput2023genretrieval, tay2022transformer} has demonstrated considerable efficacy in sequential recommendation tasks. GenRetrieval introduces items or queries as tokenized ID sequences enabling an autoregressive model to sequentially predict relevant items based on preceding context. However, fine-tuning large language models (LLMs) for this specialized purpose introduces a critical challenge known as catastrophic forgetting. This phenomenon, where a model loses previously learned information after training on a new task, causes a significant degradation of the LLM's pre-existing general-purpose capabilities. Such a trade-off between desired task-specific performance and general task competence limits the broader applicability of these models. This limitation necessitates a shift towards unified models that can concurrently excel at specialized recommendation tasks and maintain their foundational language and reasoning abilities. Mitigating catastrophic forgetting is, therefore, a central research priority.

\begin{figure}[t!]
  \centering
  \includegraphics[width=0.8\columnwidth]{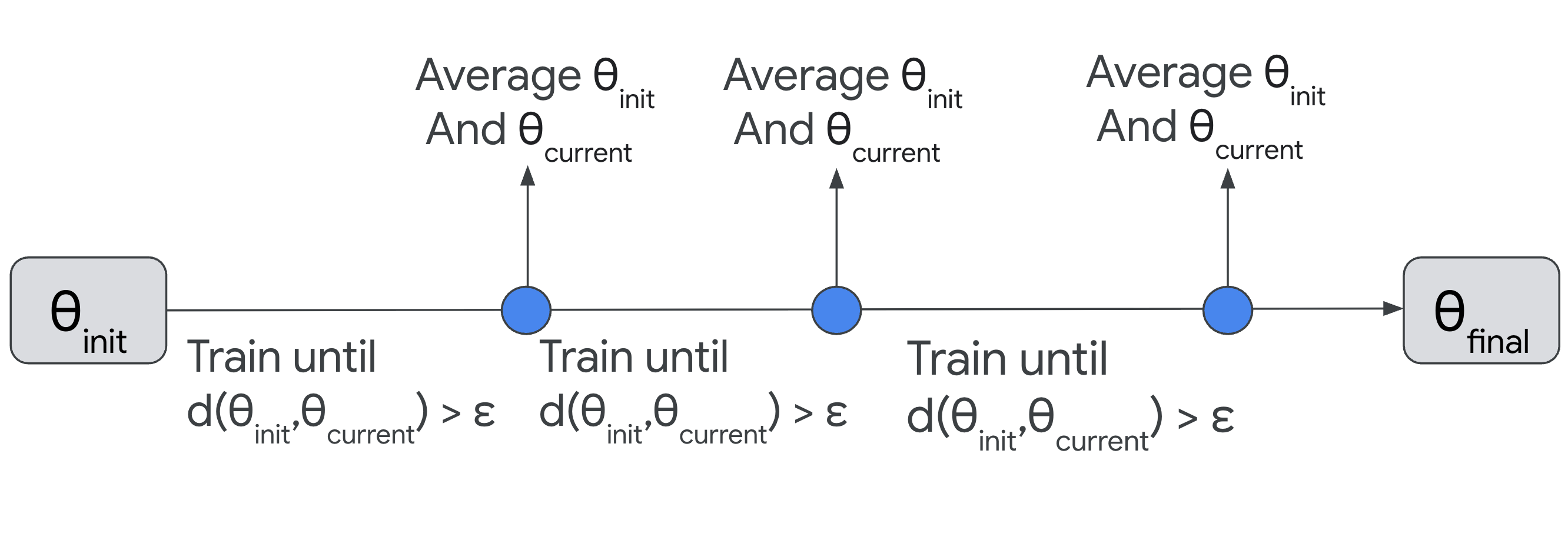}
  \caption{An overview of our \method{} method. During the fine-tuning of the downstream task, inter-model distance is tracked; when this distance exceeds a threshold $\epsilon$, weight averaging is used as a regularization step to reduce the forgetting of parametric knowledge from $\theta_\text{init}$. }
  \label{fig:overview}
\end{figure}

Due to the proprietary nature of LLM training data and the large cost of re-training a conversational agent, we focus on lightweight adaptation methods that can be applied atop a previously trained LLM while mitigating forgetting.
For this reason, we turn to model merging based methods, which are characterized by 1) their prior success in combining the capabilities of several models directly in parameter space \citep{yang2024model} and 2) their relatively lightweight nature requiring no substantial retraining.
Prior work has used model merging in continual learning settings, where merging serves as an adaptable method to combine prior and current models without inducing substantial forgetting \citep{marouf2024weighted, dziadzio2025merge, kleiman2025soup}.

In investigating the severity of the forgetting problem in GenRetrieval, we find that general LLM text-based reasoning performance is lost early and rapidly during fine-tuning. 
Relatedly, we find that post-hoc model merging methods designed to boost original model performance \textit{after fine-tuning} fail to generalize in our setting \citep{wortsman2022robust}. 
As a result, we focus on model merging techniques that apply merging steps \textit{throughout fine-tuning}. 
Motivated by the insufficiencies of one-round post-hoc model merging and by severe forgetting observed in our GenRetrieval setting, we propose \method: \textbf{O}rigin-\textbf{R}egulated \textbf{B}ack-merging of \textbf{I}ntermediate \textbf{T}rajectories.
To prevent severe forgetting, \method{} regulates the total allowable distance between original and fine-tuned model parameters during training, according to specific metrics. 
At any point when fine-tuned weights are deemed too far, original model parameters are averaged with current fine-tuned parameters. 

Compared to other regularization techniques, including methods that also employ merging during fine-tuning, we find that \method{} is Pareto-dominant as measured by recommendation performance and performance on several language-based benchmarks. 
The key contributions of this work are summarized below:
\begin{enumerate}
\item We propose \method, a method that mitigates catastrophic forgetting by tracking inter-model parameter distance and applying weight averaging to constrain drift from the original model.
\item We demonstrate that \method{} outperforms existing regularization techniques on the GenRetrieval task across multiple task datasets and across multiple text benchmarks.
\item We analyze \method{} and show that it adopts a distinct averaging schedule from fixed-length repeated merging techniques, which reflects its flexibility and adaptability to different learning behaviors. 
\end{enumerate}

\section{Related Work}
Model merging refers to a set of techniques that combine the capabilities of two or more models by combining their parameters directly in weight space. 
\citet{wortsman2022robust} introduce a simple method to reduce the forgetting in a pre-trained model after fine-tuning by simply post-hoc interpolating the pre-trained and fine-tuned models. 
A similar method, LiNeS, is also applied once after fine-tuning, and recombines the 
task vector resulting from fine-tuning with the pre-trained parameters after rescaling the task vector in a layer-wise manner \citep{wang2025lines}.
Several techniques introduced in prior work repeatedly apply model merging throughout the training or fine-tuning process.
\citet{sanyal2023early} focus only on pretraining a single model, but show that training LLMs from scratch with a high learning rate and intermittent checkpoint averaging can improve generalization.
\citet{alexandrov2024mitigating} mitigate forgetting in a multilingual setting by training branched models on different languages interleaved with merging steps. 
Other work focuses on model merging as a tool to enable continual learning; many prior techniques propose to use model merging to combine models after each task fine-tuning stage in a domain incremental learning setting \citep{marczak2024magmax, marouf2024weighted, cheng2025dam, dziadzio2025merge}.
\citet{sokar2025continual} extend this type of approach to continual learning across fine-tuning tasks learned with LoRA adapters \citep{hulora}.
While these prior methods generally apply merging methods after a model has been completely fine-tuned on a new domain, \citet{kleiman2025soup} take a slightly different approach by performing model merging sooner after a fixed number of fine-tuning steps. 
They also consider mitigating forgetting in a single-task adaptation setting, and show their fixed-length averaging scheme also generalizes to this setting.
In our work, we consider a single-task adaptation setting, but go beyond fixed-length averaging schedules by merging according to inter-model distance.

\section{Background and Motivation}

\paragraph{GenRetrieval task} 
In a GenRetrieval recommendation system, the sequential recommendation task is converted to an autoregressive generation problem via framing a user's item history as a context, and predicting the next item as its completion.
Prior work has proposed numerous ways of converting items into token-based IDs, including unstructured and naively structured IDs \citep{tay2022transformer}, semantically-motivated IDs \citep{rajput2023genretrieval}, and learned IDs \citep{sun2023learning}; in this work, we focus on the Semantic ID approach for encoding items as proposed in \citet{rajput2023genretrieval}. 
In brief, this framework uses a Sentence-T5 model to encode item features \citep{ni2022sentence}, and then uses an RQ-VAE model to quantize the embedding of the item \citep{zeghidour2021soundstream}. 

Whereas \citet{rajput2023genretrieval} uses an encoder-decoder based model to then learn the GenRetrieval task, but in this work, we adapt pre-trained LLMs for this task. 
In order to achieve this, we append the Semantic ID token vocabulary to the input and output vocabulary projections of an LLM, and include these new parameters during GenRetrieval fine-tuning.

\begin{table*}[t] %
\centering
\caption{Text datasets used for evaluating language and reasoning capabilities.}
\label{tab:eval_config}

\small %
\setlength{\tabcolsep}{4pt} %

\begin{tabular}{lcccccccc}
\toprule
& \textbf{BBH} & \textbf{GSM8K} & \textbf{MMLU-Pro} & \textbf{Drop} & \textbf{TriviaQA} & \textbf{HellaSwag} & \textbf{BoolQ} & \textbf{ARC-C} \\
\midrule
\textbf{Eval}   & sampling & sampling & scoring & sampling & sampling & scoring & scoring & scoring \\
\textbf{Metric} & Acc.     & Acc.     & Acc.    & Token-F1 & Acc.     & Acc.    & Acc.    & Acc.    \\
\textbf{N-shot} & few-shot & 8-shot   & 5-shot  & 1-shot   & 5-shot   & 10-shot & 0-shot  & 25-shot \\
\textbf{CoT}    & Yes      & Yes      & Yes     & No       & No       & No      & No      & No      \\
\bottomrule
\end{tabular}
\end{table*}

\begin{figure}[t]
    \centering
    \begin{subfigure}{0.48\textwidth}
        \centering
        \includegraphics[width=\linewidth]{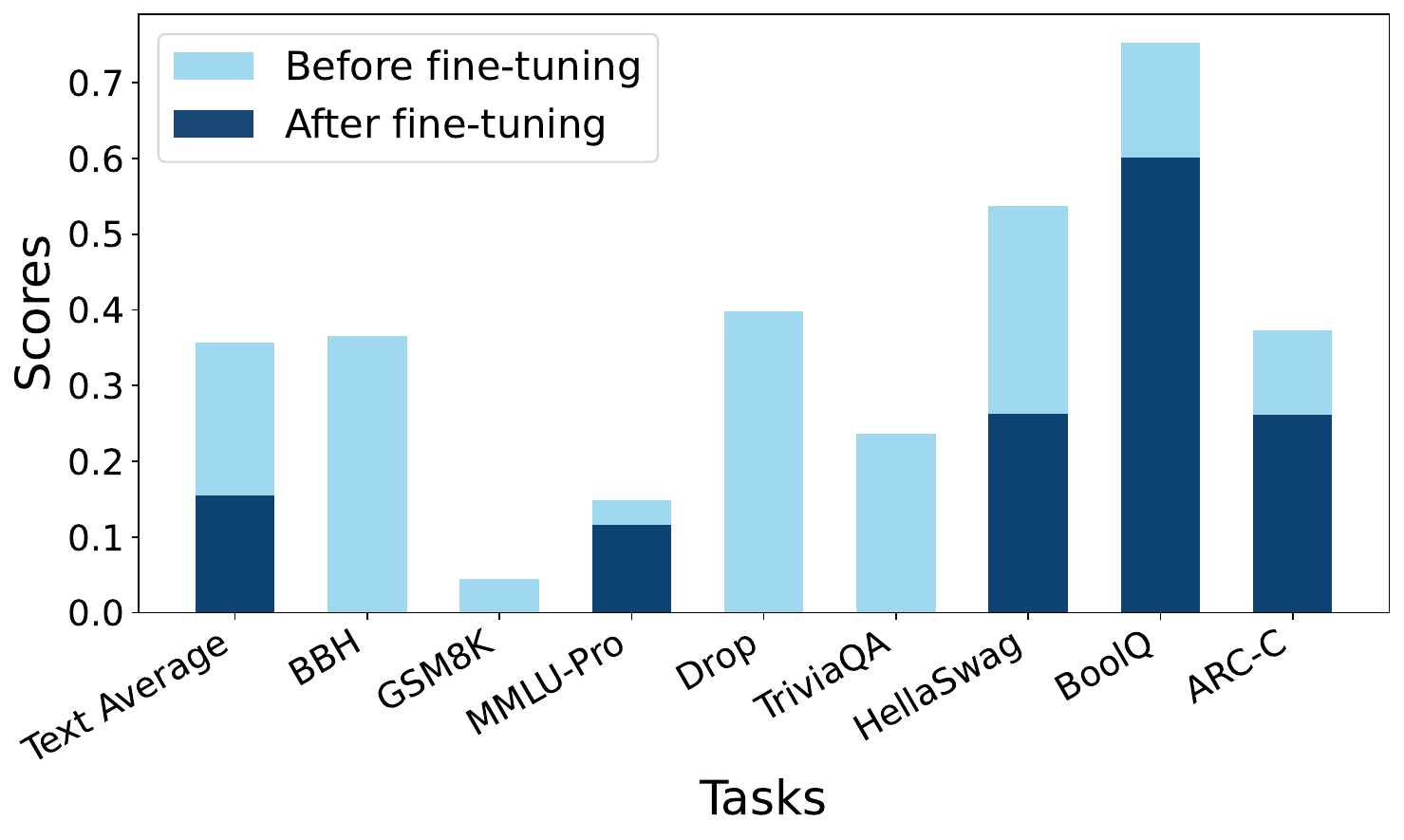}
        \caption{Text benchmark evaluations before and after fine-tuning Gemma3-1B on the Amazon Review Sports and Outdoors dataset using GenRetrieval. After fine-tuning, performance drops significantly.}
        \label{fig:quantify_forgetting}
    \end{subfigure}%
    \hfill%
    \begin{subfigure}{0.48\textwidth}
        \centering
        \includegraphics[width=\linewidth]{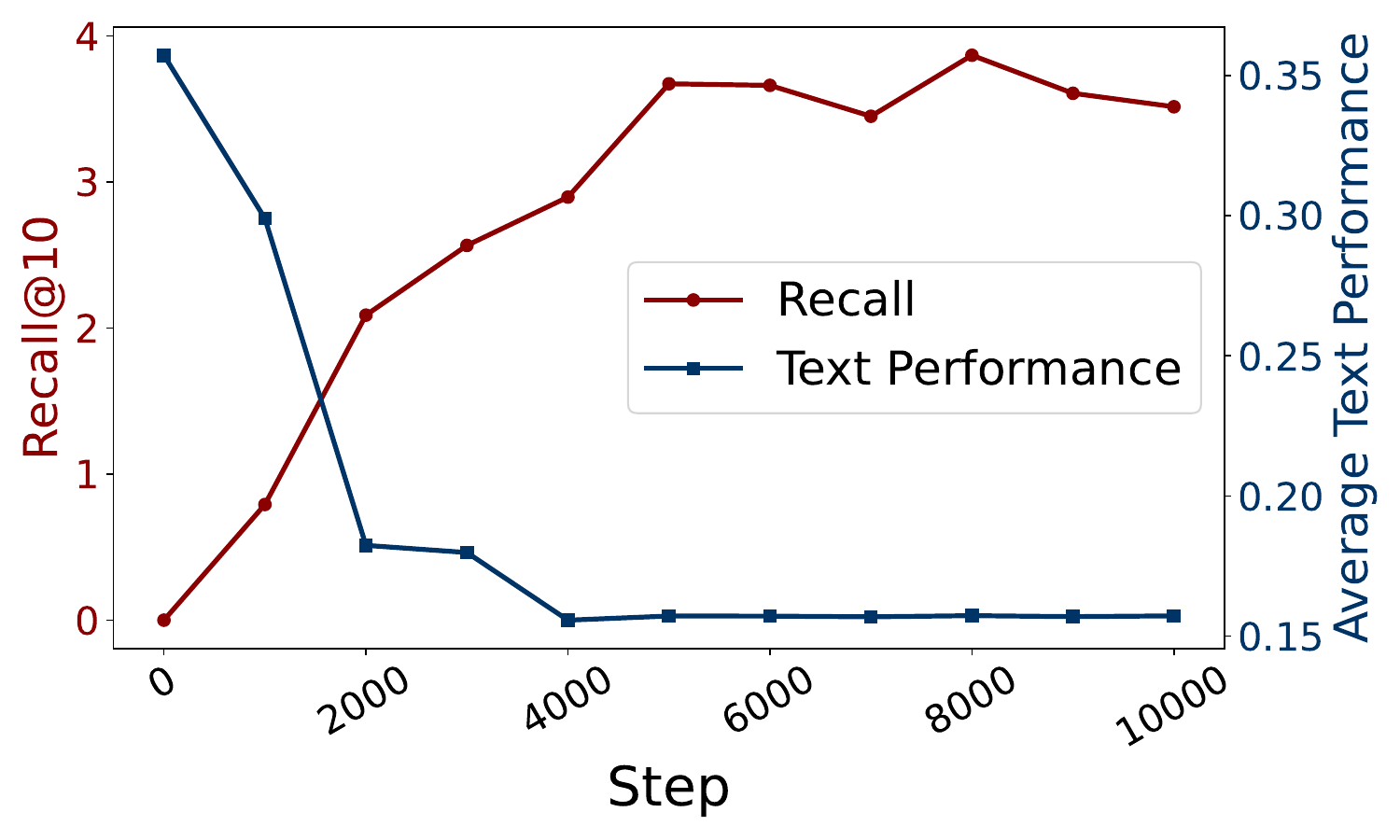}
        \caption{Recall@10 on the Amazon Review Toys and Games dataset and average text performance across our 8 benchmarks during the first 10k steps of baseline GenRetrieval fine-tuning.}
        \label{fig:rapid_forgetting}
    \end{subfigure}
    \caption{Quantitative analysis measuring forgetting during GenRetrieval finetuning.}
    \label{fig:forgetting}
    \vspace{-1em}
\end{figure}

\paragraph{Quantifying the forgetting problem in GenRetrieval}

While fine-tuning LLMs for GenRetrieval, we noticed the forgetting of original LLM capabilities. 
To demonstrate this forgetting problem, we compare an instruction-tuned Gemma3-1B model to its GenRetrieval fine-tuned counterpart on text benchmarks~\citep{team2025gemma}. 
For these exemplar experiments, we use the Amazon Product Reviews dataset \citep{he2016ups}, and report recall on the next item prediction task. 
Benchmarks to measure original LLM capability can be found in Table~\ref{tab:eval_config}. As seen in Figure \ref{fig:quantify_forgetting}, all text benchmarks drop to levels similar to random chance or majority class performance after GenRetrieval fine-tuning; for example, sampling-based benchmarks BBH, Drop, and TriviaQA drop to 0, and scoring benchmarks ARC-C and BoolQ drop to their majority class performance.

Next, we dissect this behavior more closely in our baseline GenRetrieval models to determine when forgetting occurs during fine-tuning. 
Figure~\ref{fig:rapid_forgetting} showcases the speed of forgetting during fine-tuning. Within the first 2000 steps of fine-tuning, essentially all text performance is lost, as a $\sim$0.15 benchmark average reflects a full loss of text performance. 
This suggests that we need to tailor our mitigation approach towards methods that are well suited for severe and rapid forgetting scenarios.

\begin{wrapfigure}{r}{0.5\textwidth}
\vspace{-15pt}
  \begin{center}
    \includegraphics[width=\linewidth]{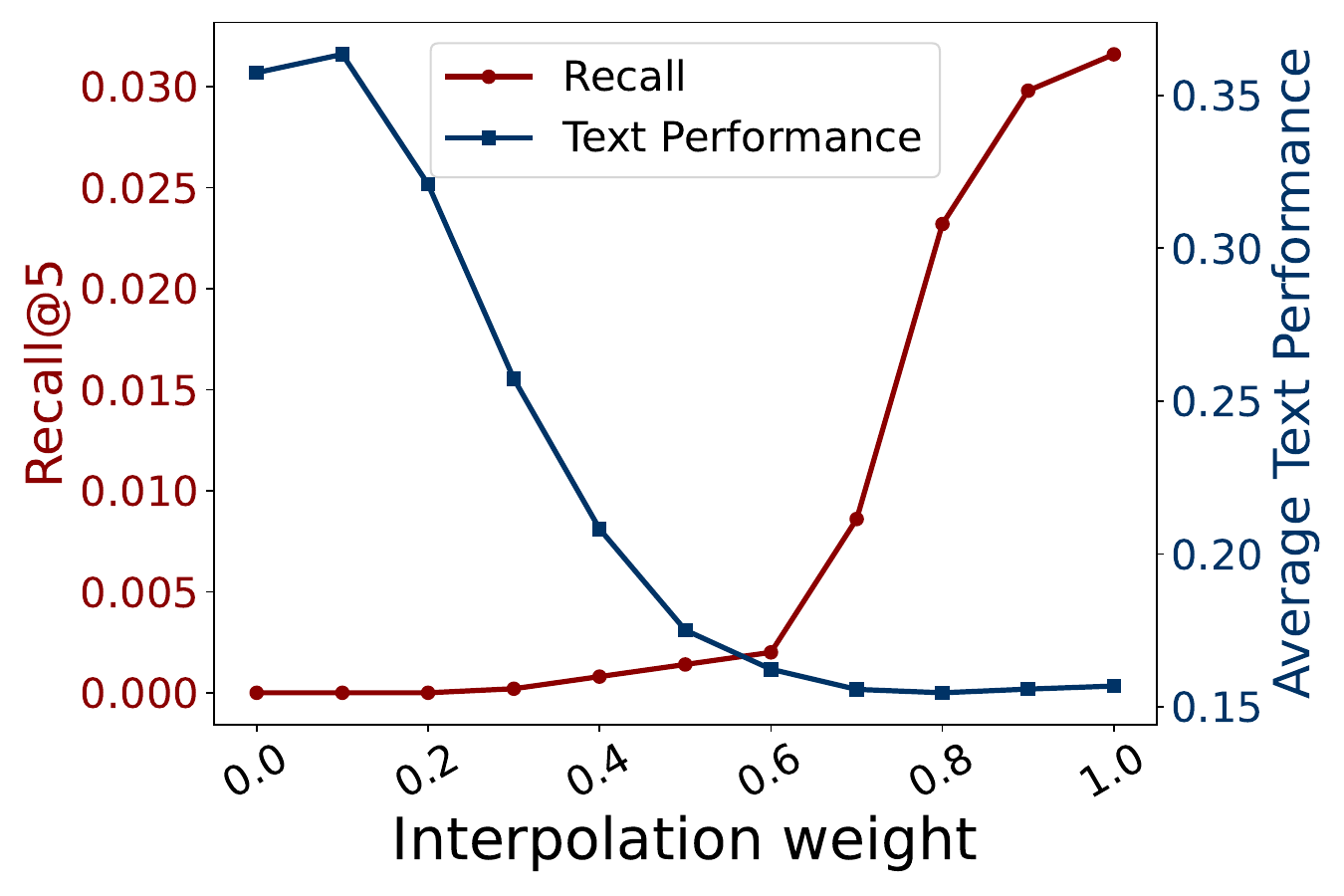}
  \end{center}
  \caption{Average text accuracy and Recall@5 performance across post-hoc, one-round weight interpolations.}
  \label{fig:posthoc}
  \vspace{-4pt}
\end{wrapfigure}
\paragraph{One-round merging fails to generalize}
To attempt the re-introduction of general capabilities back into our GenRetrieval models, we experiment with post-hoc weight interpolation between GenRetrieval weights and pretrained LLM weights to improve robustness \citep{wortsman2022robust, frankle2020linear}. 
We interpolate the final GenRetrieval and pretrained model parameters with varying interpolation ratios $\lambda$ ($\lambda=0$ reflects the initial LLM, and $\lambda=1$ reflects the GenRetrieval model), and report our results on text and recall in Figure \ref{fig:posthoc}.
As seen by the lack of simultaneous text and recall performance across interpolation ratios, simply averaging the models post fine-tuning is unable to recover both sufficient text and retrieval performance, across interpolation weights.
This failure to generalize is likely due to the rapidity of forgetting, as previously observed in Figure \ref{fig:rapid_forgetting}.
At the time of averaging, enough forgetting has likely already occurred where eventual weight averaging becomes futile.
As a result, we turn to techniques that intervene well before fine-tuning is complete.

\section{\method}
\label{sec:method}
In this section, we first define preliminaries necessary to define our method, followed by a definition of our method and observations that motivate its design.

\subsection{Preliminaries}

\paragraph{Notation} $\theta_{\text{init}} \in \mathbb{R}^p$ refers to initial LLM parameters before GenRetrieval fine-tuning. $\theta_{\text{current}} \in \mathbb{R}^p$ refers to parameters at the current fine-tuning step. 
$d$ is a distance function between two sets of parameters, $\epsilon$ is a scalar denoting a maximum distance, and $T$ is the total number of training steps. 
\paragraph{Model distance} We use two inter-model distance measures in our work. 
The first is L2-distance, which is defined as $||\theta_\text{init} - \theta_\text{current}||_2$. 
The second is Sign Dissimilarity (SD), defined as:
\begin{equation}
    \text{SD} = \frac{\text{\# of corresponding parameters with differing signs}}{\text{\# of total model parameters}}
\end{equation}
This distance is inspired in part from a simplification of metrics that measure the ``mergeability'' of models in prior model merging work \citep{yadav2023ties, sung2023empirical}.
This metric captures the fraction of parameters that have undergone a meaningful change, as measured by sign flipping, a property that is efficient to compute via bitwise XOR. 
This also means that only sign bits need to be stored for computation.
Since SID vocabulary parameters are randomly initialized for fine-tuning in $\theta_{\text{init}}$, we exclude them from distance computations.

\subsection{Algorithm}
Given the early and rapid degradation of text performance during fine-tuning, we are interested in a method that intervenes quickly during this process. 
While other methods that also employ merging during fine-tuning may be able to intervene quickly, these methods generally use a fixed cadence throughout training, which may not be the optimal schedule for the entire training duration \citep{kleiman2025soup}. Given these observations, we propose \method: Origin-Regulated Back-Merging of Intermediate Trajectories.

Our method is simple yet surprisingly effective: for a given metric, we fix a maximum distance allowable between the original model parameters, which serves as our origin, and the model parameters. 
When a training step causes the current model parameters to exceed this distance, we average the original parameters with the offending current model parameters (just after the gradient update), which we refer to as \textit{``back-merging''}.
This method schedules averaging as a function of inter-model distance, providing two major benefits. The first is its distance guarantee, where the trained model is regulated to be within a fixed distance of the original model. 
The second is the added flexibility of the averaging schedule. 
This flexibility allows for distance to dictate the averaging schedule, allowing the method to trigger merges as needed, which in turn can help improve generalizability. In brief, \method~regularizes fine-tuning via tracking a distance between $\theta_{\text{current}}$ and $\theta_{\text{init}}$, and triggering a weight averaging step if this distance is deemed too large.
We summarize \method{} in Algorithm \ref{alg:orbit}.

\begin{algorithm}
\caption{\method: Origin-Regulated Back-merging of iterative trajectories}
\label{alg:orbit}
\begin{algorithmic}[1]
    \STATE \textbf{Input:} $\theta_{\text{init}}, T, \epsilon, d(\cdot, \cdot)$
    \FOR{$t$ in ${1,...,T}$}
        \STATE \hspace{0.5cm}$\theta_{t+1}^{*} = \theta_{t} - \eta \nabla_{\theta_t} L_\text{task}$  \COMMENT{Optimizer Update}
        \STATE \hspace{0.5cm}\textbf{while} $d(\theta_{t+1}^{*}, \theta_{\text{init}}) > \epsilon$ \textbf{then}
        \STATE \hspace{0.5cm}\quad\quad $\theta_{t+1}^{*}  = \frac{\theta_{t+1}^{*} + \theta_{\text{init}}}{2}$ \COMMENT{Back-merging}
        \STATE \hspace{0.5cm} $\theta_{t+1}  = \theta_{t+1}^{*}$
    \ENDFOR
        \STATE \textbf{return} $\theta_T$
\end{algorithmic}
\end{algorithm}

A potential concern with using Sign Dissimilarity (SD) as the trigger metric in
\method{} is that unlike $L_2$ distance, SD does not contract uniformly with an arithmetic averaging step.
A sign flip at coordinate $i$ survives averaging when $|\theta_{\text{current}}(i)| > |\theta_{\text{init}}(i)|$. We show that this does not affect our within-distance guarantee in \method.\footnote{Despite allowing repeated merges, we note empirically this is never needed in our experiments. }
Crucially, if averaging fails to bring SD below the threshold $\epsilon$, the next post-update check re-triggers another merge. We demonstrate that the resulting sequence of consecutive merges drives SD to zero in a bounded number of steps.
This suffices to show that SD can be driven below some threshold $\epsilon$ within a bounded number of steps.  Our proof is in Appendix \ref{app:sd-proof}.

\subsection{Why use inter-model distance?}
\begin{figure}[t!] %
  \centering
  \begin{minipage}{0.48\textwidth}
    \centering
    \includegraphics[width=\linewidth]{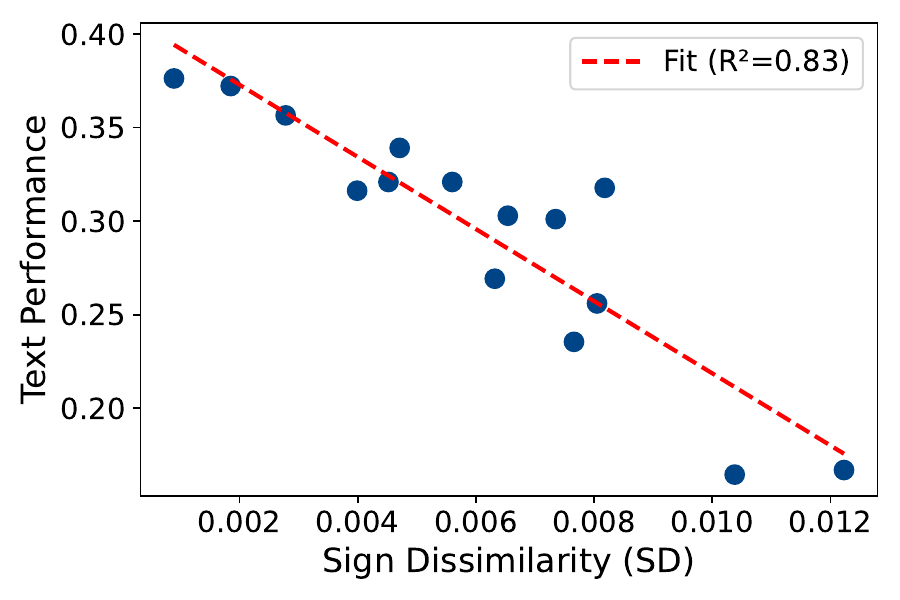}
    \caption{A scatter plot demonstrating the correlation between sign dissimilarity (SD) and average text performance. Points are collected from a Soup-to-Go experiment with a cadence of 1000 steps.}
    \label{fig:corr_text_sd}
  \end{minipage}%
  \hfill 
  \begin{minipage}{0.48\textwidth}
    \centering
    \includegraphics[width=\linewidth]{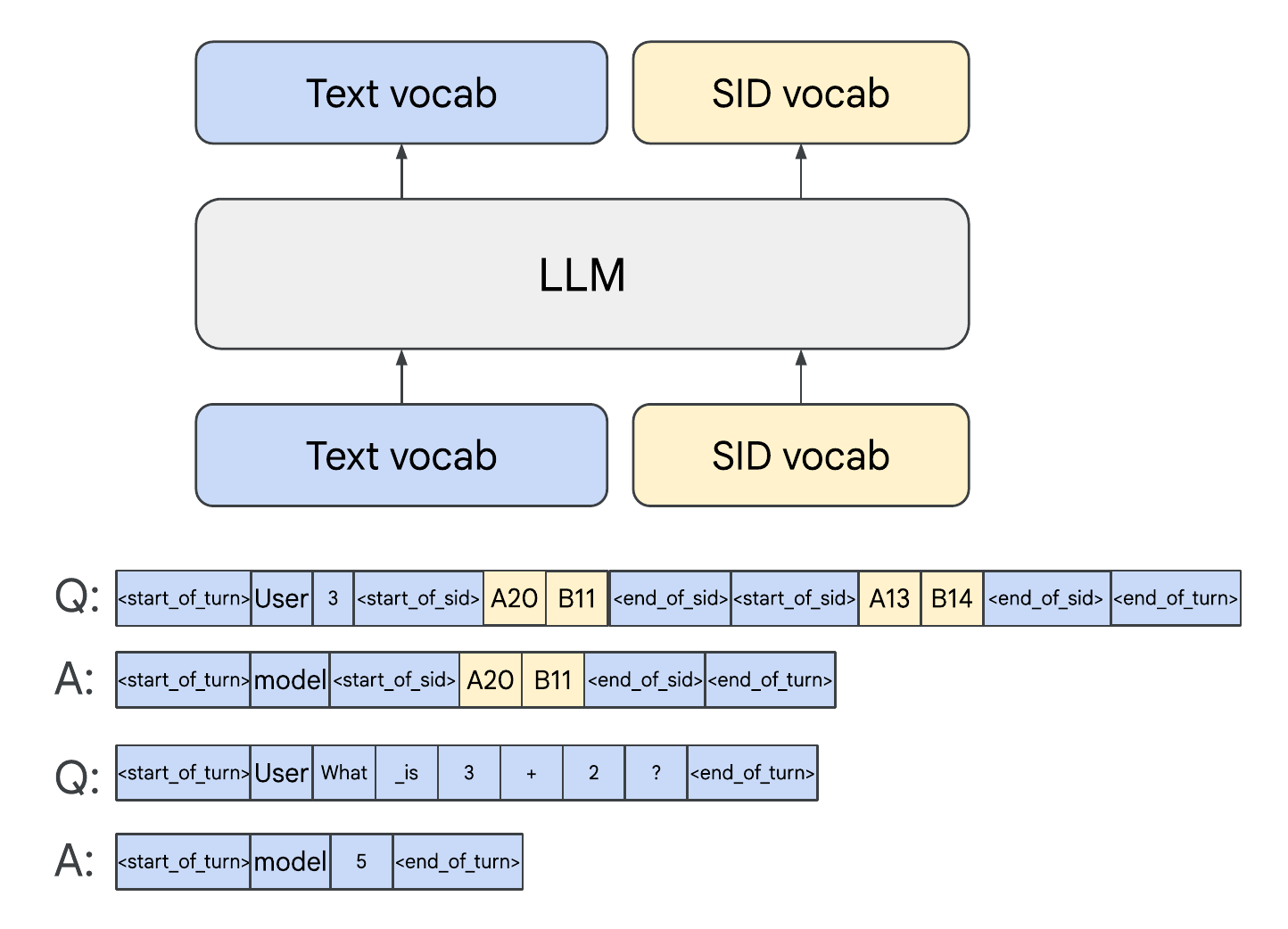}
    \caption{LLM-based GenRetrieval setup with separate text and SID vocabularies. Such a model can handle GenRetrieval queries with both SID tokens and text control tokens, and general LLM queries composed of solely text tokens.}
    \label{fig:gemma_with_index}
  \end{minipage}
\end{figure}

While averaging as a regularization tool is inspired by prior work \citep{kleiman2025soup, marczak2024magmax, marouf2024weighted}, the use of inter-model distance to schedule averaging steps is an important and novel distinction in our work.
Our motivation to use distance to dictate averaging steps comes from 1) a longstanding notion of model knowledge corresponding to localities in weight space \citep{gueta2023knowledge, wortsman2022model} and 2) a preliminary study of different checkpoints produced by Soup-to-Go fine-tuning.
Soup-to-Go is a regularization method where for a fixed number of steps $T$, the authors define a hyperparameter $0 < p < 1$ where averaging occurs every $pT$ steps between $\theta_{\text{init}}$ and $\theta_{\text{current}}$ \citep{kleiman2025soup}.
In this analysis, we choose Sign Dissimilarity as our distance measure, and we compute the text performance of these checkpoints.
We display the results of this analysis in Figure \ref{fig:corr_text_sd}, where we evaluate several checkpoints (every 2000 steps) from fine-tuning Gemma1B IT on the Amazon Product Reviews Sports and Outdoors dataset using GenRetrieval. We use Soup-to-Go to regularize this fine-tuning by averaging every 1000 steps. Additional hyperparameters are in Table \ref{tab:hyperparameters_s2g} in Appendix \ref{app:s2g_explore_hparams}. 

In testing the performance of Soup-to-Go in our setting, we observe that across checkpoints saved throughout training, there is a correlation between text performance and distance from the initial parameters. This observation suggests that fine-tuning that pushes the model starkly away from its starting point can induce severe forgetting. As a result, we focus on limiting this distance in our proposed method in order to preserve capability from the original model. 
In this setting, inter-model distance serves as a lightweight, gradient-free, and data-free proxy to forgetting.

\section{Experimental Setup}
\label{sec:experimental_setup}

\subsection{GenRetrieval Fine-tuning}
For fine-tuning our GenRetrieval models, we use with Gemma3~\citep{team2025gemma} as our base model.\footnote{Gemma3 is released under the Gemma Terms of Use.}
We use the instruction-tuned version as it more closely matches the capabilities we are interested in preserving versus the pre-trained model.
For our recommendation data, we use the Amazon Product Reviews dataset, which comprises three different subsets: Beauty, Sports and Outdoors, and Toys and Games \citep{he2016ups}. 
Dataset statistics are in Appendix \ref{app:dataset_stats}. 

\begin{wraptable}{r}{0.45\textwidth}
    \centering
    \vspace{-10pt} %
    \caption{Base GenRetrieval Hyperparameters}
    \label{tab:hyperparameters}
    \small %
    \begin{tabular}{lc}
        \toprule
        \textbf{Hyperparameter} & \textbf{Value} \\
        \midrule
        Optimizer & Adafactor \\
        LR Schedule & Cosine Decay \\
        Peak Learning Rate & 0.02 \\
        Min. Learning Rate & 1e-5 \\
        Warmup Steps & 10,000 \\
        Decay Steps & 30,000 \\
        Training Steps & 50,000 \\
        Batch Size & 16 \\
        \bottomrule
    \end{tabular}
    \vspace{-10pt} %
\end{wraptable}

We follow \citet{rajput2023genretrieval} for dataset preprocessing steps, including filtering users with less than 5 reviews, and limiting the number of items in a user’s history to 20. 
After preprocessing, each datapoint consists of a user ID, the user’s previous items encoded with Semantic IDs (SIDs), and a held-out additional item which is also converted to its SIDs. Each item uses 4 SID tokens to create the SID sequence.

We add \texttt{$<$start\_of\_SID$>$} and \texttt{$<$end\_of\_SID$>$} control tokens to the set of tokens in our base LLM, and append these tokens to each prompt in order to elicit the SID response during evaluation. To train the baseline GenRetrieval models, we logically separate SID vocabulary items from the original text vocabulary in order to compute cross-entropy loss from both text and SID vocabularies. SID parameters are also excluded from back-merging. An example of our model and an example GenRetrieval query can be found in Figure \ref{fig:gemma_with_index}. Note that for GenRetrieval finetuning, the text cross-entropy is not used beyond control tokens since the decoded tokens only contain the SID tokens. During inference, we default to text vocab and only use SID vocab for tokens between \texttt{$<$start\_of\_SID$>$} and \texttt{$<$end\_of\_SID$>$} control tokens.

To compute the possible SID sequences for computing retrieval metrics, we use beam search with 20 beams, and 20 tokens per beam. To evaluate sequential recommendation performance, we report Normalized Discounted Cumulative Gain@10 (NDCG@10) and Recall@10 values. We summarize key hyperparameters for training our baseline GenRetrieval models in Table \ref{tab:hyperparameters}. We use a cosine decay learning rate with warmup for these baselines.

\subsection{Baselines}
\paragraph{Simple baselines} We include no-intervention baselines, as well as L2-weight decay to reflect a baseline from traditional continual learning literature. 
Other techniques in this space, like Elastic Weight Consolidation and data replay methods, require data from the original model training data to reintroduce into the training mixture or compute Fisher information \citep{kirkpatrick2017overcoming}.
In our setting, we assume no access to prior training data, as we are interested in preserving the capability of LLMs trained on proprietary data. 
(1) \textbf{No interventions} refer to baseline models without any training interventions. The first of these models is simply Gemma3-1B-IT, which represents full text capability, and no GenRetrieval performance. The second baseline is the fine-tuned GenRetrieval model.
(2) \textbf{L2 weight decay} is a classic continual learning technique that adds a loss penalty to minimize the distance between the current and initial model weights. Our proposed \method{} method is similar in spirit to weight decay, but sets a strict boundary for model distance and averages parameters accordingly rather than learning according to a penalty.

\paragraph{Soup-to-Go} Soup-to-Go is a simple method designed for continual learning in deep-learning models \citep{kleiman2025soup}. While fine-tuning a model on a new domain dataset the original model weights are averaged with the current model weights after every $pT$ steps in order to preserve capabilities from the original model, where $0 < p < 1$, and $T$ is the total number of training steps. The authors generally use $<$ 10 merges during fine-tuning for their experiments, with $p > 0.1$. 
In this work, we specify Soup-to-Go baselines by their cadence with $k$ steps.

\subsection{\method{} Settings}

For both Soup-to-Go and \method, we utilize a constant learning rate during fine-tuning. While a cosine decay schedule helps the GenRetrieval baseline method improve recall performance, a constant learning rate simplifies the effect of model averaging in Soup-to-Go and \method, so different learning segments between merges are not subject to different learning rates. In turn, we fine-tune these models longer, for up to 200k steps. Longer training was not found to be helpful in our baseline GenRetrieval models. We use a learning rate of 0.001 for all temporal averaging based experiments. We test $\epsilon = 7e-3, 7.5e-3$ given the results in Figure \ref{fig:corr_text_sd}.

\subsection{Text evaluation}
\label{sec:text_eval}

To evaluate the language capabilities of our GenRetrieval models, we use the 
benchmarks from Table \ref{tab:eval_config}, as described with their evaluation settings. These benchmarks reflect a mix of scoring and sampling-based evaluations in order to cover a breadth of capabilities.
We report the average.

\section{Results}

We evaluate our baselines and \method{} for GenRetrieval fine-tuning across two primary objectives: text-based reasoning and sequential recommendation. We report average text performance, NDCG@10, and Recall@10.
To identify techniques that perform well in both domains, we employ Pareto efficiency principles.

Because Pareto-optimal sets often contain multiple solutions, we introduce a Distance To Ideal Point (DTIP) metric to select a single, balanced model. 
DTIP measures the normalized Euclidean distance between a model's performance as represented by a tuple, and an "ideal" point representing the theoretical maximum in both domains.
For a given model, let $T$ represent text performance, and $R$ represent retrieval performance (Recall@5). We first normalize the scores using the min-max scaling to a range $[0,1]$. 
\begin{equation}
    T' = \frac{T - T_{min}}{T_{max} - T_{min}} \text{~~and~~} R' = \frac{R - R_{min}}{R_{max} - R_{min}}
\end{equation}
Minimum text performance is set by the text performance observed during full, no-intervention GenRetrieval fine-tuning. Maximum performance is set by the specialized, single-domain models.
The DTIP is then calculated as the $L_2$ distance from the normalized performance $(T', R')$ to the ideal point, which is $(1,1)$, reflecting maximum text and retrieval performance. 
\begin{equation}
    DTIP = ||(1,1) - (T', R')||_2
\end{equation}
\vspace{-2em}

\begin{table}[t!]
    \centering
    \caption{Full results across Amazon Product Reviews test sets for baseline methods and \method. To identify a single representative checkpoint from a Pareto-optimal set, we measure the fraction of original performance retained for both text and retrieval. We then select the checkpoint that maximizes this combined retention. \textbf{Bold} is best performance, \underline{underline} is second best. }
    \label{tab:main_results_table}
    \resizebox{\textwidth}{!}{%
    \begin{tabular}{l *{4}{c c c}}
        \toprule
         & \multicolumn{4}{c}{\textbf{Sports and Outdoors}} & \multicolumn{4}{c}{\textbf{Toys and Games}} & \multicolumn{4}{c}{\textbf{Beauty Subsets}} \\
        \cmidrule(lr){2-5} \cmidrule(lr){6-9} \cmidrule(lr){10-13}
        \textbf{Method} & \makecell{Avg Text\\Perf.} & \makecell{NDCG\\@10} & \makecell{Recall\\@10} & \makecell{DTIP$\downarrow$} & \makecell{Avg Text\\Perf.} & \makecell{NDCG\\@10} & \makecell{Recall\\@10} & \makecell{DTIP$\downarrow$} & \makecell{Avg Text\\Perf.} & \makecell{NDCG\\@10} & \makecell{Recall\\@10} & \makecell{DTIP$\downarrow$} \\
        \midrule
        Text Baseline & 35.72 & 0 & 0 & 1.00 & 35.72 & 0 & 0 & 1.00 & 35.72 & 0 & 0 & 1.00 \\
        Retrieval Baseline & 15.52 & 2.16 & 3.76 & 1.00 & 15.59 & 3.98 & 6.57 & 1.00 & 15.69 & 3.47 & 6.06 & 1.00\\
        \midrule
        L2 Decay$_{\lambda=1e-4}$ & 15.76 & 1.92 & 3.53 & 0.99 & 15.66  & 3.67 & 6.09 & 1.00 & 15.75 & 3.17 & 5.49 & 1.00\\
        Soup-to-Go$_{k=3K}$ & 26.73 & 1.20 & 2.32 & 0.63 & 20.34 & 2.42 & 4.26 & 0.86 & 32.70 & 0.79 & 1.63 & 0.78 \\
        Soup-to-Go$_{k=2K}$ & 30.27 & 1.11 & 2.15 & \underline{0.54} & 28.55 & 2.54 & 4.36 & 0.50 & 26.44 & 1.88 & 3.64 & 0.63 \\
        \midrule
        \method$_{SD=7.5e-3}$ & 28.95 & 1.37 & 2.58 & \textbf{0.49} & 30.86 & 2.64 & 4.55 & \underline{0.41} & 25.33 & 2.28 & 4.37 & \underline{0.62} \\
        \method$_{SD=7e-3}$ & 28.88 & 1.22 & 2.42 & 0.55 & 30.98 & 2.59 & 4.40 & \textbf{0.40} & 30.47 & 2.11 & 3.90 & \textbf{0.47}\\
        \bottomrule
    \end{tabular}
    }
\end{table}

\subsection{\method{} maximizes joint performance} 

We measure the performance across the LLM benchmarks from Section \ref{sec:text_eval} and on the sequential recommendation task for each chosen subset of the Amazon Product Reviews dataset. For both \method{} and Soup-to-Go fine-tuning, we save a set of checkpoints during training that are Pareto-optimal. 
To select a single representative checkpoint from this Pareto-optimal set, we measure the DTIP, and then select the checkpoint that maximizes this combined retention.

We compare the performance of \method{} to several baselines and report our results in Table \ref{tab:main_results_table}. 
We find that \method{} improves joint performance further than Soup-to-Go, minimizing the distance to the ideal point. 
Beyond comparisons to baselines, we note that performance of a $\sim$0.3 text average on text tasks and Recall@5 of $\sim$.02 reflects substantial text and retrieval performance, compared to their topline baselines. 
We also display the set of Pareto-optimal points for these methods in Figure \ref{fig:orbit_pareto}, specifically for the Sports and Outdoors validation subset. 
As seen in the figure, \method{} improves joint performance even further than Soup-to-Go, with all four displayed variations outperforming all baselines. 
We note that in comparing all Pareto optimal checkpoints per training run, we see that \textit{every} checkpoint generated by \method{} is superior to \textit{every} checkpoint generated by baselines, which expands upon the results found in the table. 
We hypothesize that the imposition of model distance to determine averaging helps our method find solutions that maintain high text performance by design, while optimizing the GenRetrieval objective subject to this soft constraint.

\subsection{Temporal averaging techniques mitigate text forgetting in GenRetrieval}

\begin{figure}[t!] %
  \centering
  \begin{minipage}[t]{0.48\textwidth}
    \centering
    \includegraphics[width=\linewidth]{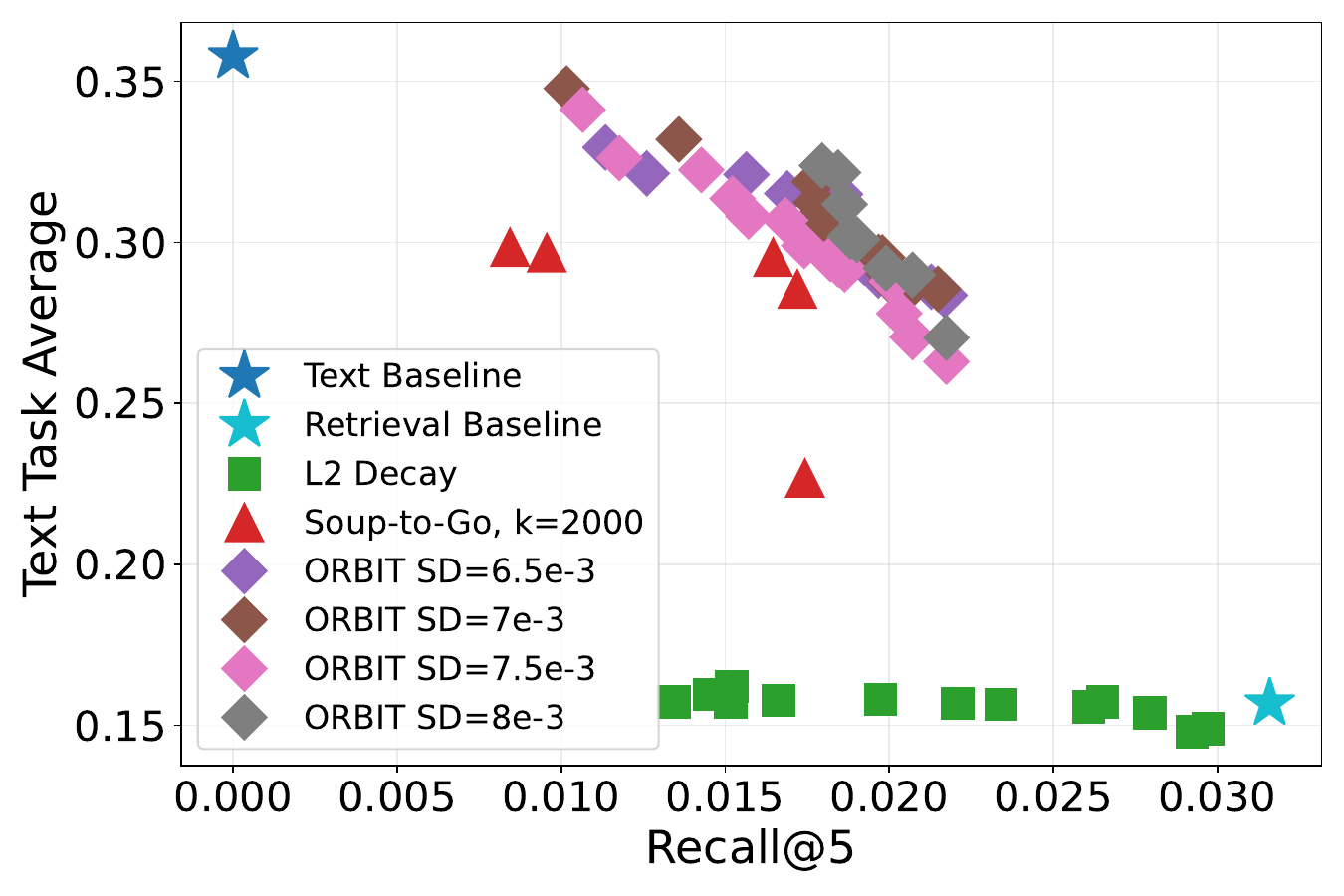}
    \caption{Text and recall performance for \method{} models, compared to a Soup-to-Go baseline and L2 decay baselines on the Sports and Outdoors dataset (validation) and our 8 text benchmarks. We display only Pareto-optimal checkpoints generated within each experiment. We can observe that all \method{} checkpoints outperform those generated from Soup-to-Go training.}
    \label{fig:orbit_pareto}
  \end{minipage}%
  \hfill 
  \begin{minipage}[t]{0.48\textwidth}
    \centering
    \includegraphics[width=\linewidth]{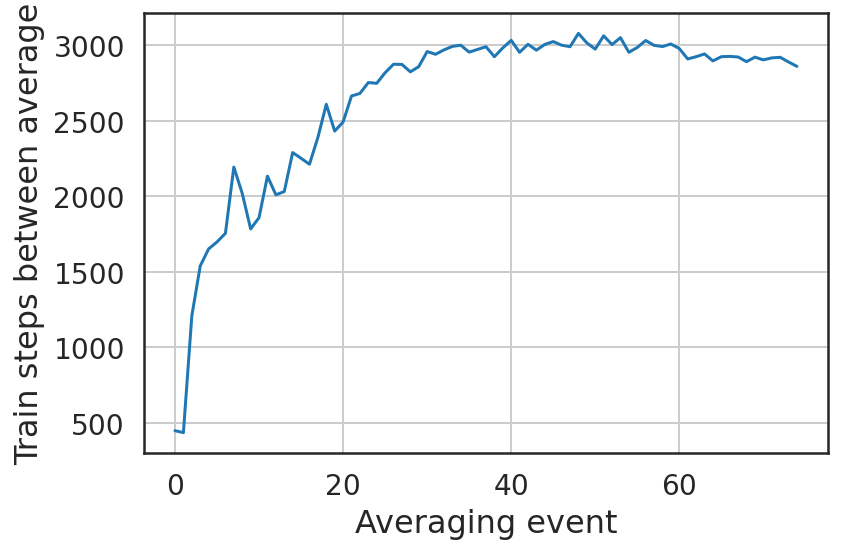}
    \caption{The number of steps between averaging events (indexed) in \method~with a maximum sign dissimilarity of 0.007, over 200k training steps. The number of steps between averages increases during training, before settling around 3000 steps later in training.}
    \label{fig:orbit_sched}
  \end{minipage}

\end{figure}

As seen in both Figure \ref{fig:orbit_pareto} and Table \ref{tab:main_results_table}, only the methods that use repeated averaging, namely Soup-to-Go and \method, are able to achieve non-trivial text and retrieval performance.
This result emphasizes the utility of regularization techniques that employ weight averaging multiple times during training; in the face of severe forgetting, classical techniques like weight decay, as well as more modern techniques like post-hoc weight averaging, may fail to generalize. 

For Soup-to-Go results, we select $k=2000, 3000$ as they provides an adequate balance between text and retrieval performance. However, we note that this value, which corresponds to $p=0.01, 0.015$ according to the original definition of the method, is much smaller than the values reported in the original paper, which are all larger than $0.15$.
This finding highlights the severe degree of forgetting, and the greater importance of repeated averaging strategies in fine-tuning LLMs for GenRetrieval. 

\section{Analysis}

\subsection{Our \method{} schedule is distinct from a constant schedule}

In designing \method{}, we hypothesized that the flexibility in the learned schedule, as determined by inter-model distance, may help prevent forgetting compared to a fixed-length averaging schedule. 
To observe the learned schedule, we compute the number of training steps between each averaging step for \method{} applied to GenRetrieval with the Sports and Outdoors dataset.

As seen in Figure \ref{fig:orbit_sched}, the number of steps between averaging steps increases over the training interval, before appearing to converge to about 3000 steps in the latter half of training. 
This “learned” schedule is distinct from what is proposed in the original Soup-to-Go work, where averaging occurs at a regular interval, which would be reflected as a horizontal line in this graph. 
This added flexibility inherent in our method may help its generalization as well to other settings. 

\subsection{Choice of Metric}

\begin{table}[ht!]
    \centering
    \begin{minipage}[t]{0.48\textwidth}
        \centering
        \caption{Metric analysis of SD and L2 in \method{}. We find that SD is preferred to L2 distance for \method{}.}
        \label{tab:metric_test}
        \resizebox{\linewidth}{!}{%
        \begin{tabular}{lcccc}
        \toprule
        \textbf{Setting} & \makecell{Avg Text\\Perf.} & \makecell{NDCG\\@10} & \makecell{Recall\\@10}  & \makecell{DTIP$\downarrow$}\\
        \midrule
        \method$_{SD=0.007}$ & 28.88 & 1.22 & 2.42 & \textbf{0.55} \\
        \method$_{L2=5}$ & 35.40 & 0.25 & 0.36 & 0.88 \\
        \method$_{L2=50}$ & 35.66 & 0.71 & 1.26 & 0.67 \\
        \method$_{L2=500}$ & 15.82 & 1.36 & 2.58 & 1.05 \\
        \bottomrule
        \end{tabular}}
    \end{minipage}%
    \hfill%
    \begin{minipage}[t]{0.48\textwidth}
        \centering
        \caption{\method{} and Soup-to-Go applied to GenRetrieval fine-tuning, using Gemma3-4B IT as the base model.}
        \label{tab:results_scaled_4b}
        \resizebox{\linewidth}{!}{%
        \begin{tabular}{lcccc}
        \toprule
        \textbf{Setting} & \makecell{Avg Text\\Perf.} & \makecell{NDCG\\@10} & \makecell{Recall\\@10} & DTIP $\downarrow$ \\
        \midrule
        Gemma3-4B IT & 57.02 & 0 & 0 & 1.00 \\
        Base GenRetrieval & 15.56 & 2.04 & 3.63 & 1.00 \\
        Soup-to-Go$_{k=2K}$ & 48.12 & 1.32 & 2.54 & 0.43\\
        \method$_{SD=0.007}$ & 47.94 & 1.35 & 2.58 & \textbf{0.42}\\
        \bottomrule
        \end{tabular}}
    \end{minipage}
\end{table}

We initially select Sign Dissimilarity as the metric in \method{} as it provides some indication of how many parameters change meaningfully, versus a combined magnitude of change, as is the case with L2 distance. 
However, we are interested in the sensitivity of \method{} to different choices of distance metrics. 
To this end, we compare the use of SD and L2 in \method{}; for SD, we fix the distance to be 0.007 given its success in prior experiments, and for L2, we test values of ${5, 50, 500}$. 
We display our result on Sports and Outdoors in Table \ref{tab:metric_test}.

We find that SD is preferred to L2 distance, but that L2 distance may also serve as a suitable metric. 
However, we recommend the use of SD given its simplicity to compute, as well as its fractional representation of parameter change.

\subsection{Scaling}

While we focus on experimenting with Gemma3-1B models, we are interested in evaluating the performance of our method on a larger model to determine the method's extensibility. 
We evaluate both Soup-to-Go and \method{} on Gemma-3-4B models, on the Sports and Outdoors dataset. 
We select hyperparameters for both methods given their best performance at the 1B scale. 
We display our results on Gemma3-4B IT in Table \ref{tab:results_scaled_4b}.
We find that performance scales with a larger model, which provides improved overall performance.

\section{Conclusion}

In this work, we introduce \method, a new method that enables language models to perform both general text and GenRetrieval functionalities. 
We design \method{} to prevent LLMs from losing their general language skills while being fine-tuned for the specialized GenRetrieval task. 
We demonstrate that \method{} preserves a meaningful amount of both recommendation and text-based reasoning capabilities in LLMs adapted for the GenRetrieval task, which in turn can help enable a unified model for conversational based discovery for recommendation systems.
We also demonstrate that compared to both classic regularization methods and related averaging-based methods, \method{} outperforms these methods by maintaining higher Pareto performance across both desired capabilities. 
While our work focuses on mitigating forgetting specific to GenRetrieval, our method is general in its construction and could extend to other applications where severe forgetting is observed.

\bibliography{example_paper}
\bibliographystyle{icml2025}

\newpage
\appendix

\section{Limitations}
\label{app:limitations}

This work focuses on mitigating forgetting in a full fine-tuning setting, common in model post-training pipelines. It does not consider parameter efficient fine-tuning (PEFT) techniques such as LoRA, which may have different forgetting behaviors during fine-tuning \citep{hulora}. 
Additionally, we test our method on 1B and 4B models, which are much smaller than frontier models. Testing whether \method{} achieves the same forgetting reduction at this scale remains necessary.

\section{Amazon Product Reviews Dataset}
\label{app:dataset_stats}
\begin{table}[htbp]
\centering
\caption{Dataset Statistics for the three subsections of the Amazon Product Reviews data.}
\label{tab:amazon_reviews_data}
\begin{tabular}{lcc}
\toprule
\textbf{Dataset} & \textbf{Users} & \textbf{Items} \\
\midrule
Beauty & 22,363 & 12,101 \\
Sports and Outdoors & 35,598 & 18,357 \\
Toys and Games & 19,412 & 11,924 \\
\bottomrule
\end{tabular}
\end{table}

\section{Soup-to-Go exploratory hyperparameters}
\label{app:s2g_explore_hparams}
\begin{table}[h]
    \centering
    \vspace{-10pt} %
    \caption{Soup-to-Go Hyperparameters in Distance Study}
    \label{tab:hyperparameters_s2g}
    \small %
    \begin{tabular}{lc}
        \toprule
        \textbf{Hyperparameter} & \textbf{Value} \\
        \midrule
        Optimizer & Adafactor \\
        LR Schedule & Cosine Decay \\
        Peak Learning Rate & 0.02 \\
        Min. Learning Rate & 1e-5 \\
        Warmup Steps & 10,000 \\
        Decay Steps & 20,000 \\
        Training Steps & 30,000 \\
        Batch Size & 16 \\
        \bottomrule
    \end{tabular}
    \vspace{-10pt} %
\end{table}

\section{Finite-Merge Recovery Guarantee for ORBIT under Sign Dissimilarity}
\label{app:sd-proof}

\subsection{Setup}

Let $\theta_{\text{init}} \in \mathbb{R}^p$ denote the origin parameters and let
$\theta_0 \in \mathbb{R}^p$ denote the current parameters during fine-tuning where
\textsc{Orbit} triggers a merge. 

Define the merge operator
\begin{equation}
M(\theta) \;=\; \tfrac{1}{2}\bigl(\theta + \theta_{\text{init}}\bigr),
\end{equation}
and let $\theta_k = M^k(\theta_0)$ denote the parameters obtained by applying $k$
consecutive merges. 

We make one mild non-degeneracy assumption: $\theta_{\text{init}}(i) \neq 0$ for
every coordinate $i$ counted in SD. 
The randomly-initialized SID vocabulary entries are explicitly excluded from the
SD computation, as discussed in Section~4.1, and pretrained weights are
generically nonzero.

\subsection{Closed form for iterated merging}

\label{lem:closed-form}
For all $k \geq 0$,
\begin{equation}
\theta_k \;=\; \frac{1}{2^k}\,\theta_0 \;+\; \bigl(1 - \frac{1}{2^k}\bigr)\,\theta_{\text{init}}.
\end{equation}

\begin{proof}
By induction on $k$. The base case $k=0$ is immediate. Assuming the formula
holds at step $k$,
\begin{align}
\theta_{k+1}
&= \frac{1}{2}\bigl(\theta_k + \theta_{\text{init}}\bigr) \\
&= \frac{1}{2}\Bigl(\frac{1}{2^k}\theta_0 + (1 - \frac{1}{2^k})\theta_{\text{init}} + \theta_{\text{init}}\Bigr) \\
&= \frac{1}{2^{k+1}}\theta_0 + \bigl(1 - \frac{1}{2^{k+1}}\bigr)\theta_{\text{init}}. \qedhere
\end{align}
\end{proof}

\subsection{Per-coordinate sign flip}

\label{lem:per-coord}
For a coordinate $i$ with $\theta_{\text{init}}(i) \neq 0$, and suppose
$\mathrm{sign}(\theta_{0}(i)) \neq \mathrm{sign}(\theta_{\text{init}}(i))$.
Define the magnitude ratio
\begin{equation}
r_i \;=\; \frac{|\theta_{0}(i)|}{|\theta_{\text{init}}(i)|}.
\end{equation}
Then sign flipping after $k$ merges will occur, $\mathrm{sign}(\theta_{k}(i))= \mathrm{sign}(\theta_{\text{init}}(i))$, 
if and only if
\begin{equation}
2^k \;>\; 1 + r_i.
\end{equation}

\begin{proof}
By Result \ref{lem:closed-form}, $\theta_{k}(i) = \frac{1}{2^k}\theta_{0}(i) +
(1 - \frac{1}{2^k})\theta_{\text{init}}(i)$. Without loss of generality assume
$\theta_{\text{init}}(i) > 0$. By the assumption, $\theta_{0}(i) < 0$. Then $\theta_{k}(i) > 0$ iff
\begin{align}
\frac{1}{2^k}\theta_{0}(i) + (1 - \frac{1}{2^k})\theta_{\text{init}}(i) &> 0 \\
 (2^k - 1)\,\theta_{\text{init}}(i) &> -\theta_{0}(i) \\
\iff\; 2^k &> 1 + r_i. \qedhere
\end{align}
\end{proof}

\noindent Coordinates that are not flipped at $\theta_0$ remain aligned
with $\theta_{\text{init}}$ for every $k \geq 0$, since averaging two same-sign
values cannot change the sign. 

\subsection{Finite-merge recovery}

\label{thm:finite-merge}

Let $\mathcal{F} = \{\,i : \mathrm{sign}(\theta_{0}(i)) \neq
\mathrm{sign}(\theta_{\text{init}}(i))\,\}$ denote the set of initially-flipped
coordinates, and let $r_{\max} = \max_{i \in \mathcal{F}} r_i$. 
For any $k >  \log_2(1 + r_{\max}) $,
\begin{equation}
\mathrm{SD}(\theta_k, \theta_{\text{init}}) \;=\; 0.
\end{equation}

\begin{proof}
By Result~\ref{lem:per-coord}, the set of coordinates contributing to
$\mathrm{SD}(\theta_k, \theta_{\text{init}})$ at merge step $k$ is exactly
\begin{equation}
\mathcal{F}_k \;=\; \bigl\{\,i \in \mathcal{F} : r_i \geq 2^k - 1\,\bigr\},
\end{equation}
which is monotonically nonincreasing in $k$.

Choose $k$ so that $2^k - 1 > r_{\max}$, i.e.,
$k > \log_2(1 + r_{\max})$. Then $\mathcal{F}_k = \emptyset$
and $\mathrm{SD}(\theta_k, \theta_{\text{init}}) = 0$.

\end{proof}

\newpage

\end{document}